\documentclass{article}
\usepackage{iclr2027_conference,times}

\usepackage{amsmath,amsfonts,bm}

\def\eqref#1{equation~\ref{#1}}

\def\1{\bm{1}}

\def\rc{{\textnormal{c}}}

\DeclareMathAlphabet{\mathsfit}{\encodingdefault}{\sfdefault}{m}{sl}
\SetMathAlphabet{\mathsfit}{bold}{\encodingdefault}{\sfdefault}{bx}{n}

\newcommand{\E}{\mathbb{E}}

\usepackage[utf8]{inputenc}
\usepackage[T1]{fontenc}
\usepackage{xcolor}
\usepackage{hyperref}
\usepackage{url}
\usepackage{amsmath}
\usepackage{amssymb}
\usepackage{booktabs}
\usepackage{pifont}
\usepackage{float}
\usepackage{graphicx}
\usepackage{multirow}
\usepackage{xspace}
\usepackage{microtype}
\usepackage{placeins}

\newcommand{\method}{\textbf{Periscope}\xspace}

\title{\vspace{-1.5em}Periscope: Extending Frozen Language Models Beyond Their Context Window}

\author{%
\makebox[\textwidth][c]{\textbf{Mohamed Eltahir$^{1*}$}\hspace{2.5em}\textbf{Anas Obayd$^{1*}$}\hspace{2.5em}\textbf{Raed Rashid$^{1}$}\hspace{2.5em}\textbf{Abdulrahman Alghamdi$^{1}$}}\\[0.25em]
\makebox[\textwidth][c]{\textbf{Abdulrahman Mousa$^{1}$}\hspace{2.5em}\textbf{Abdallah Ahmed$^{1}$}\hspace{2.5em}\textbf{Tanveer Hussain$^{2\dagger}$}\hspace{2.5em}\textbf{Naeemullah Khan$^{1\ddagger}$}}\\[1.0em]
\makebox[\textwidth][c]{$^{1}$King Abdullah University of Science and Technology (KAUST), Thuwal, Saudi Arabia}\\
\makebox[\textwidth][c]{$^{2}$Department of Computer Science, Edge Hill University, Ormskirk, England}\\[0.3em]
\makebox[\textwidth][c]{\small\texttt{\{mohamed.hamid, anas.obayd, raed.rashid, abdulrahman.alghamdi,}}\\
\makebox[\textwidth][c]{\small\texttt{abdulrahman.mousa, abdallah.ahmed, naeemullah.khan\}@kaust.edu.sa}}\\[0.1em]
\makebox[\textwidth][c]{\small\texttt{htanveer3797@gmail.com}}
}

\iclrfinalcopy 
\begin{document}

\maketitle

 \lhead{Preprint.}

 \vspace{-2.00em}

 {\renewcommand{\thefootnote}{\fnsymbol{footnote}}
 \footnotetext[1]{Equal contribution.}
\footnotetext[2]{Corresponding author.}
 \footnotetext[3]{Principal Investigator (PI).}}
{\renewcommand{\thefootnote}{}\footnotetext{Code: \url{https://github.com/mohammad2012191/Periscope}}}

\vspace{-1.0em}

\begin{figure}[H]
\centering
\includegraphics[width=\linewidth]{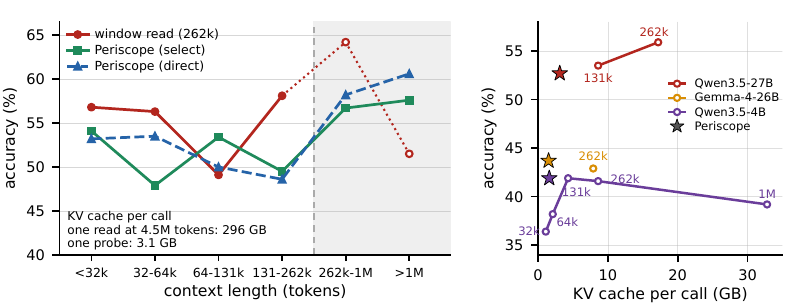}
\caption{Qwen3.5-27B on LongBench v2. \textbf{Left:} Accuracy by context length. The window read at the native 262k window is dotted past it, where it reads only a prefix. \textbf{Right:} Accuracy against the peak key-value cache of one call, three models (Table~\ref{tab:main}). Circles: the window read at each window. Stars: \method{}'s selected read.}
\label{fig:reach}
\end{figure}

\vspace{-1.5em}

\begin{abstract}
A language model reads long text in one quadratic forward pass, stops at the context window, and loses accuracy with length before reaching it. We ask whether the read can be factorized when deciding over a finite set: which document is relevant, which option is supported, which passage is the evidence. \method{}, a training-free inference method, arranges the $N$ chunks of a text on a $K{\times}K$ grid with $K{=}\lceil\sqrt{N}\rceil$ and asks a frozen model the same question about $K$ local spans of consecutive chunks and $K$ strided spans that sample the whole text, reading the log-odds of every answer at one token. Each answer takes its best local and strided score, and scoring every chunk by its two spans gives an evidence map at no further cost, whose peak is the chunk behind the answer. Every probe is about $\sqrt{sc}$ tokens for a text of $s$ tokens and chunk size $c$, so a window of $W$ tokens reaches $W^{2}/c$ tokens at $s^{1.5}$ cost. The map replaces the long read. On LongBench v2, reading only the $K$ chunks the map ranks highest, 9k tokens, matches the same model's best window read across windows from 32k to 1M tokens, and on InfiniteBench, where the median context is 150k tokens, it leads the best window read by 5 points. The same map ranks BRIGHT's long-document corpora with the best NDCG@10 of six methods. Each call caches only one probe, so a 27B model reads 4.5M-token contexts on one 80GB GPU, where a single pass would need 296GB of cache. A long read then needs a GPU that holds the model, not one that holds the text.
\end{abstract}

\section{Introduction}\label{sec:intro}

A language model reads long text in a single forward pass, and the pass has three limits. Its attention cost grows with the square of the input. It stops at the context window. And it loses accuracy before it gets there: most long-context models fall below a quality threshold well before their claimed length \citep{ruler}, 11 of 13 models claiming 128k tokens fall below half of their short-context accuracy by 32k once the needle shares little vocabulary with the question \citep{nolima}, and accuracy drops with input length even when every relevant passage is present \citep{ducontext}.

Three responses exist, each with a structural limit. Extending the window \citep{yarn, longrope, selfextend} moves the wall and keeps the quadratic cost and the monolithic memory of the pass. Reading less, by retrieving chunks with an embedding model \citep{dpr} or compressing the prompt \citep{llmlingua}, lets another model decide what enters the pass, with no guarantee that what it kept is what the question needs. Reading in pieces bounds every call, but passage scoring sees each piece alone \citep{daicallan} and a chain of agents passes a lossy summary from one piece to the next \citep{chainofagents}, so nothing relates the final answer to a read of the whole text.

We start from what the read is for. Many long-context reads are decisions over a finite set: which document is relevant, which option is supported, which passage is the evidence. Such a decision needs only one supporting passage. Mainly, a document is relevant if some parts of it support the query, and an option is correct if some passage establishes it. A frozen model can judge, from a bounded window, whether that window supports an answer, and the bounded window is also where the model is accurate. So the text can be covered by bounded windows and the answers composed by a maximum, which adds no approximation of its own, and the answer distribution over the windows is an estimate of where the evidence sits.

\begin{figure}[t]
  \centering
  \includegraphics[width=\linewidth]{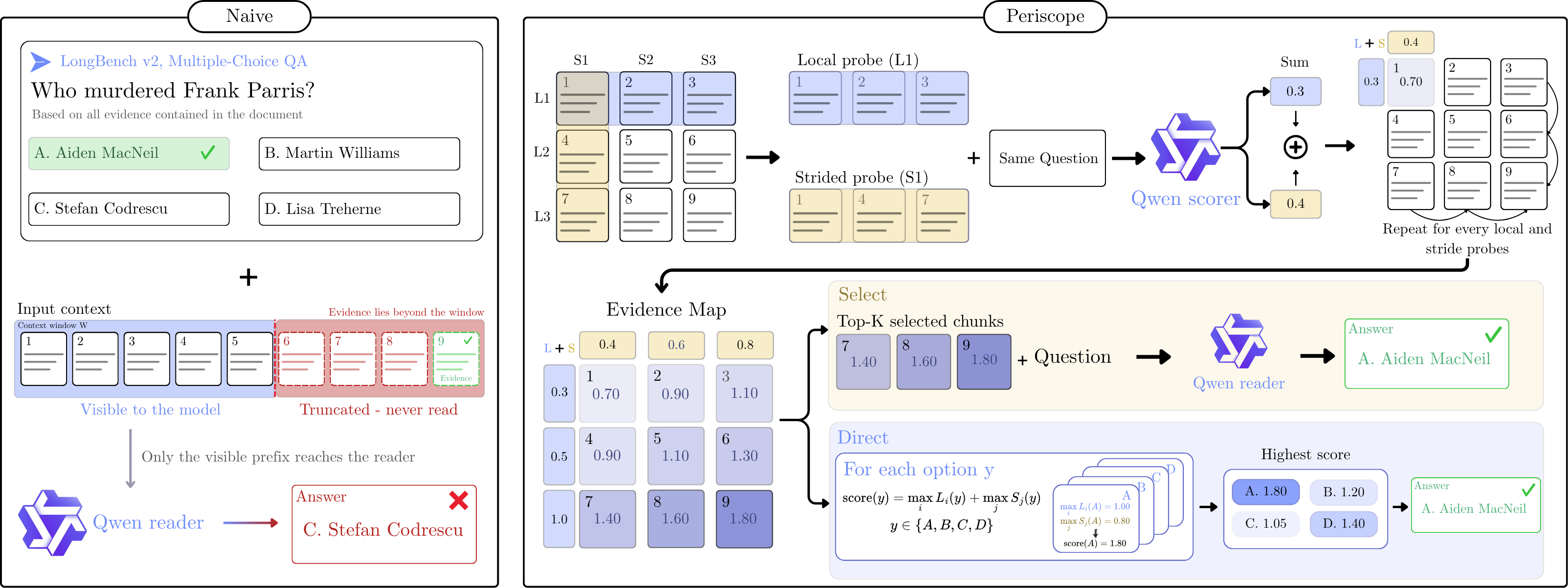}
  \caption{\method{} on a LongBench v2 question. Left: the window read sees only the first chunks and misses the evidence in chunk 9. Right: the chunks sit on a $K{\times}K$ grid, here $3{\times}3$, each local probe $L_i$ reads $K$ consecutive chunks and each strided probe $S_j$ every $K$-th chunk, and each chunk scores the sum of its two probes. The map peaks at the evidence chunk, and the answer comes from reading the top $K$ chunks (Select) or from each option's best local and strided scores alone (Direct).}
  \label{fig:pipeline}
\end{figure}

\method{} (Figure~\ref{fig:pipeline}) is that read: an inference method with which a frozen model served at a window of $W$ tokens reads texts of up to $W^{2}/c$ tokens, for chunk size $c$, with memory bounded per call. It uses the same forward pass and the same one-token readout as the single read, applied to bounded inputs, with nothing generated, no state between calls, and no orchestration. The $N$ chunks of a text are arranged on a $K{\times}K$ grid with $K{=}\lceil\sqrt{N}\rceil$. The model answers the same question about $K$ \emph{local} spans, each of $K$ consecutive chunks, and $K$ \emph{strided} spans, each joining every $K$-th chunk so that it samples the whole text. Every chunk is read twice by $2K$ independent calls of about $\sqrt{sc}$ tokens for a text of $s$ tokens, at a cost growing as $s^{1.5}$. Each answer takes its best local and its best strided score, and scoring every chunk by the two spans it lies in gives a map of where the evidence sits, whose peak is the chunk behind the answer. The same map serves every task through a fixed readout: it ranks documents, answers the question directly, locates the evidence, and selects a read of the $K$ chunks it ranks highest, one probe long.

We evaluate the method as inference, holding the model, the prompt, and the readout fixed and changing only which text reaches each forward pass. On LongBench v2 \citep{longbenchv2}, reading the $K$ chunks the map ranks highest, 9k tokens, matches the same model's best window read across windows from 32k to 1M tokens and leads embedding, random, and prefix selection at the same budget by 6 to 11 points. Each call caches one probe, so the selected read matches or approaches the best window read with a third to a fifth of its cache (Figure~\ref{fig:reach}). On InfiniteBench En.MC \citep{infinitebench}, with a median context of 150k tokens, the same read leads the best window read by 5 points. On BRIGHT \citep{bright} the ranking readout attains the best NDCG@10 and MRR of six methods.

\textbf{Contributions.} (1) \emph{Factorized reading.} A decomposition that captures the context of a whole text through short, independent local and strided probes, so a model with a window of $W$ tokens reads texts of $W^{2}/c$ tokens without ever reading them at once. (2) \emph{The evidence map.} The probes' answers combine into a map of where the evidence sits in the text, at no extra cost. (3) \emph{Readouts of the map.} The same map ranks documents, answers questions directly, locates the evidence, and selects a short read that replaces reading the whole window.
\section{Related Work}\label{sec:related}

\textbf{Limits of the single long read.}
Most long-context models fall below a quality threshold well before their claimed length, and only half hold up at 32k tokens \citep{ruler}, the gap widens once the needle shares little vocabulary with the question \citep{nolima}, content in the middle of a long input is used less than content at either end \citep{lostmiddle}, and accuracy falls with input length even when retrieval is perfect \citep{ducontext}. The limit these studies share is the pass itself: every token competes with every other for the same decision, and the model is most reliable on the short inputs it reads best.

\textbf{Extending the window.}
The first response is to make the pass longer. Positional interpolation and its successors extend a pretrained window by orders of magnitude with little or no tuning \citep{pi, yarn, longrope, selfextend}, and keep the quadratic attention and the memory of the original. Sub-quadratic architectures address the cost itself. State-space and linear-attention layers carry a fixed-size state \citep{mamba, gateddeltanet}, and sparse attention reads a selected subset of keys \citep{deepseekv32}. A fixed state alone copies and retrieves from context poorly \citep{repeatafterme}, so deployed models keep a minority of full-attention layers \citep{jamba, minimax01}, whose key-value cache still grows with every token. The largest of these models need a multi-GPU node for a long read: Jamba-1.5-Large fits a 256k context on eight 80GB GPUs with 8-bit expert weights \citep{jamba15}, and MiniMax-Text-01 was sized to process over 1M tokens on a single eight-GPU, 640GB machine with 8-bit quantization \citep{minimax01}. These are also architectures rather than inference methods, trained or continually pretrained with their layers in place, so they do not apply to an existing model. In every case the whole text enters one pass, and the evaluations above find accuracy declining well inside the window.

\textbf{Reading less.}
The second response keeps the pass short by deciding in advance what enters it. Retrieval-augmented generation reads the top-$k$ chunks a retriever selects \citep{rag, dpr}, and prompt compression drops tokens before the pass \citep{llmlingua}. Dense and late-interaction retrievers \citep{dpr, colbert} score passages from representations computed before the query is seen, and such retrievers perform poorly when relevance requires reasoning rather than surface matching \citep{bright}. Memory-augmented models retrieve blocks of the reader's own key-value cache by similarity, and for EM-LLM also by temporal contiguity \citep{infllm, emllm}. In every case a selection rule fixed before or outside the reader's judgment of the answer decides what is read, and what the selector misses cannot be recovered by the read.

\textbf{Language models as judges.}
The reader can make that decision itself. Pointwise rerankers read a relevance score from a language model's one-word answer, trained \citep{monot5} or zero-shot, possibly after generated query and document analyses \citep{judgerank}, and listwise and setwise rerankers read an ordering \citep{rankgpt, setwise}, with the cost per judgment accounted in FLOPs by \citet{e2rflops}. A judgment read from a single answer token is cheap and reliable, but rerankers assume the document fits in one call and that a first stage assembled the candidates.

\textbf{Reading in pieces.}
A text that does not fit one call can be judged in parts. FirstP scores the lead passage and MaxP the best passage of a document \citep{daicallan}, so every call is bounded, but no call sees beyond its own passage. Chain of Agents processes chunks in sequence, each worker passing a generated message to the next and a manager writing the answer \citep{chainofagents}. LongAgent has members read chunks in parallel and report to a leader over several rounds, each round depending on the leader's previous state \citep{longagent}. These systems carry information across the text, but their calls depend on earlier generated outputs, each message is an intermediate that can drop what a later chunk would have needed, and the final answer has no stated relation to a read of the full text. Bounded, independent calls that still see across the whole text would need a different way to cut it.

\textbf{Separable structure and answer-space probing.}
A matrix built from one vector along its rows and one along its columns is the cheapest structure a matrix can have, and it is used wherever the full matrix is too costly to measure or learn, from separable filters to low-rank weight updates \citep{lora}. Grid probing carried this structure to inference: it places video frames on a $K{\times}K$ grid, asks a frozen model about each row and each column, and combines the answers into a map that selects frames for a second pass, $2K$ calls for $K^{2}$ frames \citep{gridprobe}. The number of calls grows with the square root of the number of items rather than with the number itself. A long text has no such geometry of its own. If it could be folded into a separable grid, the same structure might avoid the two bottlenecks of the single read, the quadratic cost of full attention and the fixed boundary of the window.

Together these threads describe what a read past the window needs: calls that are bounded and independent, a decision made by the reader itself, a view of the whole text and not only of its parts, and a score that shows where its evidence lies.

\section{\method{}}\label{sec:method}

\method{} replaces the single forward pass over a text with $2K$ bounded passes and a composition rule. The task determines only the answer set of the probe and which readout of the composed scores is used. The model, its serving, and its readout are those of the single read. Nothing is generated, no state passes between calls, and the calls are independent.

\subsection{The probe}\label{sec:probe}

A probe is one forward pass of a frozen model $\theta$ on a span $S$ of the text and the question $q$, read at the first answer token. It returns one score per answer, measuring how strongly the span alone supports that answer. For a multiple-choice question with options $\mathcal{Y}$, the prompt lists the options and one more, an abstain option $\bar{y}$. The probe score $\ell(y\mid S,q)$ of each option $y$ is its log-odds against abstaining,
\begin{equation}
\ell(y\mid S,q)\;=\;\log p_\theta(y\mid S,q)\;-\;\log p_\theta(\bar{y}\mid S,q),
\label{eq:probe}
\end{equation}
where $p_\theta(\cdot\mid S,q)$ is the model's probability of each answer at that token. The abstain option lets the model report that the span holds no evidence, so such a span scores every option low rather than favoring one at random, and scores from different spans can be compared. For relevance, the question asks whether $S$ is relevant to a query, and the score is the log-odds of Yes against No. 

\subsection{Local and strided spans}\label{sec:grid}

We split a text $x$ of $s$ tokens into $N{=}\lceil s/c\rceil$ chunks $x_1,\dots,x_N$ of $c$ tokens, set $K{=}\lceil\sqrt{N}\rceil$, pad to $K^{2}$ chunks with empty ones, and index the chunks as a $K{\times}K$ grid so that chunk $x_{(i-1)K+j}$ occupies cell $(i,j)$. The $i$-th local span and the $j$-th strided span are
\begin{equation}
\mathcal{L}_i=\bigl(x_{(i-1)K+1},\dots,x_{iK}\bigr),
\qquad
\mathcal{S}_j=\bigl(x_{j},\,x_{j+K},\dots,x_{j+(K-1)K}\bigr),
\label{eq:spans}
\end{equation}
each a concatenation of $K$ chunks in reading order. A \emph{local} probe reads $\mathcal{L}_i$, a stretch of $K$ consecutive chunks. A \emph{strided} probe reads $\mathcal{S}_j$, every $K$-th chunk, a uniform sample of the whole text. Every chunk lies in exactly one local and one strided span, so it is read twice. Each span holds $K$ chunks of $c$ tokens, about $\sqrt{sc}$ tokens, so every probe is a short call, and the $2K$ probes do not depend on each other. We write $L_i(y)$ for the score of answer $y$ from the $i$-th local probe and $S_j(y)$ for its score from the $j$-th strided probe.

\subsection{Composition, map, and readouts}\label{sec:compose}

An answer is supported by a text if some part of the text supports it, so one strong span is enough. We therefore take, for each answer, its best score in each family and add the two,
\begin{equation}
\mathrm{score}(y)\;=\;\max_i L_i(y)\;+\;\max_j S_j(y).
\label{eq:agg}
\end{equation}
An answer scores high only when some local span and some strided span both support it. 

\textbf{The map.} Each chunk lies in one local span $\mathcal{L}_i$ and one strided span $\mathcal{S}_j$, and the two scores of those spans give the chunk a score,
\begin{equation}
M[i,j]\;=\;\max_{y\neq\bar{y}}\bigl(L_i(y)+S_j(y)\bigr),
\label{eq:map}
\end{equation}
where cell $(i,j)$ is chunk $x_{(i-1)K+j}$. The map needs no further calls, and it agrees with the answer: the largest cell for $y$ equals $\mathrm{score}(y)$, so the peak of the map is the chunk whose two spans produced the answer. 

\textbf{Readouts.} The same $2K$ probes serve four tasks through four fixed readouts, none of which changes the probes.
\begin{itemize}
\item \textbf{Ranking.} With $\mathcal{Y}=\{\text{Yes},\text{No}\}$, $\mathrm{score}(\text{Yes})$ is one number per document, and a corpus is ranked by it.
\item \textbf{Direct answer.} With options as the answer set, $\arg\max_{y\neq\bar{y}}\mathrm{score}(y)$ answers the question with no further call.
\item \textbf{Localization.} The peak of $M$ names the chunk that produced the decision.
\item \textbf{Selected read.} A second stage after the probes. The $K$ chunks with the highest cells of $M$ are joined in their original order, and the model reads them with the question in one more call and answers, as a single read would. The read is $Kc$ tokens, the length of one probe, so the two stages together cost $2K+1$ probes.
\end{itemize}

\subsection{Reach, cost, and memory}\label{sec:cost}

\textbf{Reach.} We first ask how long a text a model with a window of $W$ tokens can read this way. A probe reads $K$ chunks of $c$ tokens, so it fits the window when $Kc\le W$, ignoring the short prompt. The largest grid therefore has $K\approx W/c$, and it holds $K^{2}$ chunks, $K^{2}c$ tokens in all. The longest text the window can read is
\begin{equation}
s_{\max}\;=\;K^{2}c\;\approx\;\Bigl(\frac{W}{c}\Bigr)^{2}c\;=\;\frac{W^{2}}{c}
\label{eq:reach}
\end{equation}
tokens, $W/c$ times the window itself. At $c{=}500$, a 32k window gives $K\approx65$ and reaches 2M tokens, and a 128k window gives $K\approx256$ and reaches 33M. This bounds the length a window can read, not the accuracy of the read, which Section~\ref{sec:longbench} measures.

\textbf{Cost.} We next compare the compute of the probes with that of a single read of the same text. Let $a$ be the FLOPs a model spends on each token in the layers that process tokens one at a time, which grows with its number of parameters, and $b$ the FLOPs its attention spends on each pair of tokens, which grows with its number of layers and their width. A forward pass over $t$ tokens then costs about $at+bt^{2}$ FLOPs, since attention compares every token with every other. Ignoring the input (short) prompt, a single read of the text and the $2K$ probes of $\sqrt{sc}$ tokens each cost
\begin{equation}
F_{\mathrm{full}}(s)\;=\;as+bs^{2},
\qquad
F_{\mathrm{probe}}(s)\;=\;2K\bigl(a\sqrt{sc}+b\,sc\bigr)\;=\;2as+2b\sqrt{c}\,s^{3/2}
\label{eq:fprobe}
\end{equation}
using $K=\sqrt{s/c}$. The probes do twice the per-token work, because every chunk is read twice, but their attention cost grows as $s^{1.5}$ rather than $s^{2}$. Short texts are therefore cheaper to read whole and long texts cheaper to probe, and the two costs are equal at
\begin{equation}
s^{\star}\;=\;\bigl(\sqrt{c}+\sqrt{c+a/b}\bigr)^{2}\;\approx\;a/b,
\label{eq:crossover}
\end{equation}
which is tens to hundreds of thousands of tokens for current models. 

\textbf{Memory.} A pass stores a key-value cache for every token it reads, so a single read holds the cache of all $s$ tokens, and for long texts that cache outgrows the model's weights. A probe holds the cache of its $\sqrt{sc}$ tokens only, $K=\sqrt{s/c}$ times less than the single read, and the factor grows with the text: the longer the text, the larger the saving. Because a probe never exceeds the window, its cache never exceeds that of one window, whatever the length of the text. The memory a long read needs is then set by the model's weights, not by the text, and a model that fits a GPU can read texts far longer than that GPU could cache in one pass. Section~\ref{sec:costexp} measures the compute, the latency, and the memory of both reads.
\section{Experiments}\label{sec:experiments}

\subsection{Setup}\label{sec:setup}

All models are frozen, served with vLLM in bfloat16 on one A100 80GB, and read at a single answer token with the settings of Appendix~\ref{app:probe}. Every arm runs on the same GPU, so no arm has more memory or compute than another. Every baseline is training-free. Chunks are 500 tokens for question answering and 100 for retrieval.

\subsection{Benchmarks \& Baselines}

\textbf{Question answering} uses LongBench v2 \citep{longbenchv2}, 503 four-option questions over contexts of 8k to 2M words in three splits, short (180), medium (215), and long (108), and InfiniteBench En.MC \citep{infinitebench}, 229 four-option questions over whole novels with a median context of 150k tokens. The models are Qwen3.5-4B and Qwen3.5-27B \citep{qwen35card} and the mixture-of-experts Gemma-4-26B-A4B-it \citep{gemma4}. The \textbf{window read} reads the longest prefix of the context that fits the window, and we run it at every window the GPU can serve for each model: 32k, 64k, 131k, and 262k natively and 1M with YaRN for the 4B, 131k and 262k for the 27B, and 262k for Gemma. \method{} needs far less: its longest probe is 48k tokens, so it runs at every model's native window, without extension. \method{} is run in two ways. \textbf{Direct} answers from the $2K$ probes alone, with no further call. \textbf{Select} has two stages: the probes build the map, and one more call reads the $K$ chunks the map ranks highest, about 9k tokens, and answers. Three baselines replace the map in the second stage and read the same number of chunks: the first $K$, $K$ at random, and the top $K$ by cosine similarity to the question under Qwen3-Embedding-4B. Every arm answers every question.

\textbf{Retrieval} uses the long-document corpora of BRIGHT \citep{bright}, 3,792 documents with medians of 334 to 6,701 tokens by domain and maxima above 1.8 million, and 350 queries stratified sampled across seven of BRIGHT domains, each scored against its whole domain corpus with no candidate pooling. The model is Qwen3-4B-Instruct-2507 \citep{qwen3} at a 16k window, and it scores every document with the relevance probe of Section~\ref{sec:probe}. \method{} is the ranking readout. The window read has the same model read the longest prefix of each document that fits, as a pointwise LLM reranker does \citep{judgerank}, FirstP reads the first chunk \citep{daicallan}, and BM25 \citep{bm25} and Qwen3-Embedding-4B, the parameter-matched similarity control, are run under the identical protocol. We report NDCG@10, the benchmark's headline metric, MRR, and recall at 1 and 10.

\subsection{The map replaces the read}\label{sec:longbench}

\begin{table}[t]
\centering\small
\setlength{\tabcolsep}{4.5pt}
\caption{LongBench v2 accuracy (\%) by split and the cost of each arm. \emph{Read}: tokens in the final call, $K$ chunks being 9k on average. \emph{PFLOPs}: mean compute per question from the exact token counts of every call. \emph{KV}: peak key-value cache of one call, GB. A dash: the read does not fit one A100 80GB. Bold: best per column within a block.}
\label{tab:main}
\begin{tabular}{@{} l r r r cccc @{}}
\toprule
\textbf{Method} & Read & PFLOPs & KV & Short & Medium & Long & All \\
\midrule
\multicolumn{8}{@{}l}{\emph{(a) Qwen3.5-4B}} \\
Window read, 32k & 32k & 0.3 & 1.1 & 41.1 & 34.0 & 33.3 & 36.4 \\
Window read, 64k & 64k & 0.6 & 2.1 & \textbf{45.6} & 33.5 & 35.2 & 38.2 \\
Window read, 131k & 131k & 1.2 & 4.3 & 45.0 & 38.6 & \textbf{43.5} & \textbf{41.9} \\
Window read, 262k & 262k & 2.4 & 8.6 & \textbf{45.6} & 37.2 & \textbf{43.5} & 41.6 \\
Window read, 1M (YaRN) & 1M & 8.4 & 32.8 & 43.9 & 34.4 & 40.7 & 39.2 \\
Window read, whole context & 4.5M & -- & 148 & -- & -- & -- & -- \\
First $K$ chunks & $K$ chunks & 0.1 & 1.6 & 30.0 & 29.8 & 36.1 & 31.2 \\
Random $K$ chunks & $K$ chunks & 0.1 & 1.6 & 36.7 & 30.7 & 31.5 & 33.0 \\
Embedding, $K$ chunks & $K$ chunks & 1.9 & 1.6 & 36.7 & 33.0 & 39.8 & 35.8 \\
\method{}, direct & none & 4.5 & 1.6 & 43.3 & 36.7 & 41.7 & 40.2 \\
\method{}, select & $K$ chunks & 4.6 & 1.6 & 41.1 & \textbf{41.9} & \textbf{43.5} & \textbf{41.9} \\
\midrule
\multicolumn{8}{@{}l}{\emph{(b) Qwen3.5-27B}} \\
Window read, 131k & 131k & 6.3 & 8.6 & 57.2 & 52.1 & 50.0 & 53.5 \\
Window read, 262k & 262k & 10.8 & 17.2 & \textbf{57.8} & \textbf{53.5} & \textbf{57.4} & \textbf{55.9} \\
Window read, 1M (YaRN) & 1M & -- & 65.5 & -- & -- & -- & -- \\
Window read, whole context & 4.5M & -- & 296 & -- & -- & -- & -- \\
Embedding, $K$ chunks & $K$ chunks & 2.3 & 3.1 & 46.1 & 40.5 & 54.6 & 45.5 \\
Map of the 4B, $K$ chunks & $K$ chunks & 5.1 & 3.1 & 51.1 & 45.6 & 52.8 & 49.1 \\
\method{}, direct & none & 29.7 & 3.1 & 54.4 & 48.8 & \textbf{57.4} & 52.7 \\
\method{}, select & $K$ chunks & 30.2 & 3.1 & 52.8 & \textbf{53.5} & 50.9 & 52.7 \\
\midrule
\multicolumn{8}{@{}l}{\emph{(c) Gemma-4-26B-A4B-it}} \\
Window read, 262k & 262k & 2.8 & 7.9 & \textbf{43.3} & \textbf{43.7} & 40.7 & 42.9 \\
Window read, whole context & 4.5M & -- & 136 & -- & -- & -- & -- \\
\method{}, direct & none & 4.9 & 1.5 & 41.1 & 38.6 & 45.4 & 40.9 \\
\method{}, select & $K$ chunks & 5.0 & 1.5 & 42.2 & \textbf{43.7} & \textbf{46.3} & \textbf{43.7} \\
\bottomrule
\end{tabular}
\end{table}

\textbf{A read of $K$ chunks matches the best window read.} Across the five windows we run for the 4B, its window read peaks at 131k, at 41.9 on all 503 questions. Reading only the $K$ chunks the map ranks highest, 9k tokens, scores the same 41.9, and every other window scores lower (Table~\ref{tab:main}). Against this best window, the split falls where the cost model of Section~\ref{sec:cost} puts it: on the short split every context fits and the window read is ahead by 3.9 points, on the medium split, where 44\% of contexts exceed 131k tokens, the map is ahead by 3.3, and on the long split, where all of them do, the two tie. By domain (Appendix~\ref{app:paired}, Table~\ref{tab:lbdomains}), the selected read leads or ties wherever the evidence sits in one region and trails only on domains where the evidence is spread across documents.

\textbf{Why a longer window does not help.} A longer window lets the model see more of the text, yet the 4B scores 38.2 at 64k, 41.9 at 131k, and then 41.6 at 262k and 39.2 at 1M with YaRN, where 93\% of the contexts fit whole. The 1M read drops even on the short split, whose contexts already fit at 131k, so the longer window costs accuracy without adding any text. A probe avoids it because it never sees a long input. It reads about $\sqrt{sc}$ tokens, 9k on average and 48k at the longest context, so the factorized read stays in the range where the model is accurate. Figure~\ref{fig:reach} (left) shows where this matters for the 27B, against its best window read, at the native 262k. Inside that window, where the whole context is read, the window read leads in three of four bins. Between 262k and 1M tokens it still leads, since its prefix covers at least a quarter of each context. Past 1M tokens the prefix covers less, and on these 33 questions the 262k read scores 51.5, while the map's read scores 57.6 and the direct readout 60.6. 

\textbf{The gain comes from the map.} Matching the best window read could come from the budget alone, if any $K$ chunks were enough. It does not: at the same budget, the $K$ chunks the map ranks highest score 6 points above the chunks most similar to the question, 9 above random chunks, and 11 above the first chunks, on every split. The map also picks the chunks the whole read relies on. On the 298 contexts that fit the 131k window, reading the map's chunks gives the same answer as reading the whole context on 71\% of questions, against 64\% for embedding selection and 58\% for random chunks. Even with no read at all, the 27B's direct readout tracks the whole read, its option scores correlating with the whole read's at $r{=}0.80$ (Figure~\ref{fig:fidelity}).

\textbf{A stronger model gains past the window.} The 27B fits one A100 at 131k and 262k, and not at 1M. Its best window read, at 262k, scores 55.9 on all questions, 3.2 points above either readout of \method{}, within a point of its 131k read. The two meet on the long split, where every context exceeds 131k tokens. There the direct readout, answering from the probes with no read at all, scores 57.4, the same as the 262k read and 7.4 points above the 131k read, where the 4B's direct readout does not gain. The 262k read reaches the same accuracy only by holding 17 GB of cache per call, against 3.1 for a probe.

\textbf{The result carries across models.} The map need not come from the reader. Chosen by the 4B and read by the 27B, it leads embedding selection by 4 points, so a small model can choose what a large one reads. Nor is the result specific to the hybrid attention of the Qwen models. With Gemma-4-26B-A4B-it, a mixture of experts served at its native 262k, the selected read leads the window read, 43.7 against 42.9, ties it on the medium split, and leads by 5.6 points on the long split, with a fifth of the cache.

\subsection{A second benchmark}\label{sec:infbench}

\begin{table}[t]
\centering\small
\setlength{\tabcolsep}{5pt}
\caption{InfiniteBench En.MC, Qwen3.5-4B, 229 questions, median context 150k tokens.}
\label{tab:infbench}
\begin{tabular}{@{} l r r r c @{}}
\toprule
\textbf{Method} & Read & PFLOPs & KV & Acc. \\
\midrule
Window read, 131k & 131k & 1.8 & 4.3 & 74.2 \\
Window read, 262k & 262k & 3.3 & 8.6 & 77.3 \\
First $K$ chunks & $K$ chunks & 0.1 & 0.6 & 44.1 \\
Random $K$ chunks & $K$ chunks & 0.1 & 0.6 & 42.8 \\
\method{}, direct & none & 2.9 & 0.6 & 80.8 \\
\method{}, select & $K$ chunks & 3.0 & 0.6 & \textbf{82.1} \\
\bottomrule
\end{tabular}
\end{table}

On InfiniteBench the contexts are whole novels, and two thirds of them exceed the 131k window. The map's read of $K$ chunks scores 82.1, 5 points above the best window read, at 262k, and 8 above the read at 131k (Table~\ref{tab:infbench}). The gain comes from the novels that do not fit: on the 79 that fit 131k tokens the 131k read is ahead, 82.3 against 78.5, as on the short split of LongBench v2, and on the 150 that do not the map's read leads by 14 points, 84.0 against 70.0.

\subsection{Is the map right?}\label{sec:bright}

\begin{table}[t]
\centering\small
\setlength{\tabcolsep}{4pt}
\caption{BRIGHT long-document retrieval, full corpus, 350 queries over 3,792 documents, one frozen model (Qwen3-4B) for FirstP, the window read, and \method{}. \emph{Window read}: the longest prefix that fits 16k tokens. \emph{Chunk-max}: every 512-token chunk embedded, a document scored by its best chunk. \emph{R@$k$}: recall at $k$.}
\label{tab:bright}
\begin{tabular}{@{} l cccc @{}}
\toprule
\textbf{Method} & NDCG@10 & MRR & R@1 & R@10 \\
\midrule
BM25 & .219 & .233 & .079 & .318 \\
Qwen3-Embed-4B & .461 & .450 & .213 & \textbf{.671} \\
Qwen3-Embed-4B, chunk-max & .422 & .401 & .208 & .615 \\
\midrule
FirstP & .222 & .240 & .118 & .300 \\
Window read, 16k & .453 & .453 & .225 & .633 \\
\textbf{\method{}} & \textbf{.493} & \textbf{.491} & \textbf{.266} & .666 \\
\bottomrule
\end{tabular}
\end{table}

The probe scores must also be right on their own, apart from the read that follows them. Table~\ref{tab:bright} tests the ranking readout, with no read at all, on BRIGHT, where relevance takes reasoning rather than word overlap. The probes rank the corpora with the best NDCG@10 and MRR of six methods. Against the embedder, a retriever trained for the task, the frozen model leads MRR and Recall@1, and trails by half a point on Recall@10: the verdict is sharper at the head of the ranking, while similarity fills the top ten more completely.

\subsection{Cost}\label{sec:costexp}

\begin{figure}[t]
\centering
\includegraphics[width=0.9\linewidth]{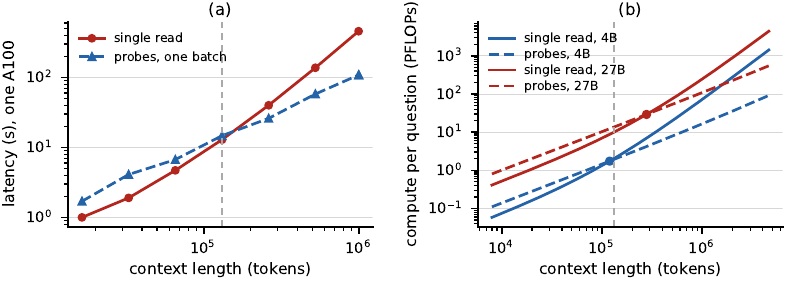}
\caption{When do the probes cost less than the read? (a) Wall-clock of one single read against one batch of its $2K$ probes, Qwen3.5-4B on one A100 served at 1M tokens. (b) Compute of both arms from Eq.~\ref{eq:fprobe} with measured coefficients, crossovers marked.}
\label{fig:cost}
\end{figure}

The probes process every token twice in calls of about $\sqrt{sc}$ tokens, so on LongBench v2 they process 2.3 and 2.5 times the tokens of the window read on the short and medium splits and 14 times on the long split. Whether that is more compute than the window read depends on the crossover of Section~\ref{sec:cost}, which is 119k tokens for the 4B and 278k for the 27B from their measured coefficients (Figure~\ref{fig:cost}(b), Appendix~\ref{app:cost}). Measured on one A100, a batch of the 4B's probes takes 26 s at 262k tokens against 40 s for the single read, and 108 s at 1M against 457 s (Figure~\ref{fig:cost}(a)). Memory does not depend on the crossover. At the longest context of the benchmark, 4.5M tokens, one probe of the 27B holds 3.1 GB of cache next to 54 GB of weights, where a single read would hold 296 GB, more than three A100s, so the 27B reads it on one GPU. (Figure~\ref{fig:reach}(right)) The 4B served with only 16 GB caches at most 178k tokens in one pass, and its single read ends there while every probe still runs. Past the window, the single read is not available at any price.

\begin{table}[t]
\centering\small
\caption{Ablations. (a) Local probes with and without the strided probes, direct readout, with BRIGHT NDCG@10 on 150 queries. (b) Independent probes against a chain over the same $K$ local spans, with latency per question, LongBench v2, Qwen3.5-4B.}
\label{tab:ablations}
\begin{minipage}[t]{0.50\linewidth}
\centering
\textbf{(a)} Local and global views\\[3pt]
\begin{tabular*}{\linewidth}{@{}l@{\extracolsep{\fill}}cc@{}}
\toprule
Probes & LongBench v2 & BRIGHT \\
\midrule
one per chunk (MaxP) & -- & .551 \\
local only & 39.4 & .551 \\
local $+$ strided (default) & \textbf{40.2} & \textbf{.637} \\
\bottomrule
\end{tabular*}
\end{minipage}\hfill
\begin{minipage}[t]{0.44\linewidth}
\centering
\textbf{(b)} Reading in pieces\\[3pt]
\begin{tabular*}{\linewidth}{@{}l@{\extracolsep{\fill}}cc@{}}
\toprule
Method & Acc. & Latency \\
\midrule
Chain of Agents & 32.0 & 207 s \\
\method{}, selected read & \textbf{41.9} & \textbf{63 s} \\
\bottomrule
\end{tabular*}
\end{minipage}
\end{table}

\section{Ablations}\label{sec:ablations}

We run two ablations with the probes held fixed: what the strided probes add, and how the probes compare with reading in sequence.

\textbf{The strided probes supply the global view.} Each local probe sees one neighborhood of the text, and each strided probe a sample of all of it, so removing the strided probes isolates what the global view adds (Table~\ref{tab:ablations}a). On BRIGHT, reading each chunk alone, as MaxP does, and reading the local spans give the same NDCG@10, .551, so finer local pieces add nothing. Adding the strided probes lifts it to .637. What the local view lacks is not resolution but a view of the whole document, and strided probes supply it with that. On LongBench v2 the strided probes add 0.8 on all questions.

\textbf{Independent probes or a chain.} Chain of Agents \citep{chainofagents}, run with the same model, and the same $K$ local spans, scores 10 points below the selected read, at three times the latency (Table~\ref{tab:ablations}b). Its $K{+}1$ calls run in sequence, and each message is written before the next span is read, so what a later span would have made relevant may already be gone. 

\section{Conclusion and Limitations}\label{sec:conclusion}

We asked whether a long read can be factorized when it is a decision. \method{} reads a text through $2K$ bounded local and strided probes of a frozen model, reaching $W^{2}/c$ tokens from a window of $W$ with the cache of one probe, while achieving a comparable performance of full read. 

\textbf{Limitations and future work.} The method covers decisions over a finite answer set. It joins evidence from distant spans only in the selected read, which trails the window read on multi-document QA, and below the crossover it costs more than a single read. Two directions follow: readouts for open-ended outputs such as summaries, and distilling the probe scores and maps into retrievers and rankers far cheaper than the model that produced them.



\section*{Acknowledgment}
We are grateful to the KAUST Academy for its generous support, and especially to Prof.\ Sultan Albarakati who made this work possible. For computer time, this research used Ibex managed by the Supercomputing Core Laboratory at King Abdullah University of Science \& Technology (KAUST) in Thuwal, Saudi Arabia.

\bibliography{bibliography}
\bibliographystyle{iclr2027_conference}

\clearpage
\appendix
\section{Appendix}
\subsection{Implementation}\label{app:probe}

\textbf{Prompts.} Every call is one user turn of the model's chat template. Both the relevance and question probes place the text first so that it is a shared prefix across queries or questions. The relevance probe is
\begin{quote}\small\texttt{Document: \{span\}\\Query: \{query\}\\Is this document relevant to the query? Answer Yes or No:}\end{quote}
with No as the abstain option. The question probe is
\begin{quote}\small\texttt{\{span\}\\\\Question: \{question\}\\A. \{option A\}\\B. \{option B\}\\C. \{option C\}\\D. \{option D\}\\E. Unsure\\Answer with a single letter:}\end{quote}
with \texttt{E. Unsure} as the abstain option $\bar{y}$. The selected read and the window read use the same prompt without the line \texttt{E. Unsure}. Each call is one forward pass with \texttt{max\_tokens=1}. We read the top-20 log-probabilities at the answer position and compute Eq.~\ref{eq:probe}. Thinking mode is disabled where the model has one.

\textbf{Sampling.} Temperature $1.0$ with \texttt{top\_p}$=1$, \texttt{top\_k}$=-1$, and \texttt{min\_p}$=0$. Temperature rescales every log-odds by the same constant and cannot change a ranking. Nucleus and top-$k$ truncation can remove an answer token from the returned distribution, so they are disabled.

\textbf{Answer tokens.} The surface form of the answer token (\texttt{Yes} against \texttt{\textvisiblespace Yes}) depends on model and template. We take the maximum over surface variants and verify at startup on one probe with a known answer.

\textbf{Chunking and grid.} Texts are tokenized once and split into $c$-token chunks. $K{=}\lceil\sqrt{N}\rceil$ and the grid is padded with empty chunks that are dropped from the spans, so no chunk is discarded, and a span that would be empty is not issued. The chunks of a span are joined by a single space with no delimiter token, so a strided span reads as running text with a discontinuity at each chunk boundary. Scores are used as read, with no calibration across spans. Chunk lists are cached per document.

\textbf{Serving.} Each model is served with vLLM on a single A100 80GB GPU, the 4B at 32 concurrent sequences and the 27B at 8, and every arm of that model, the window read included, runs against the same instance. The $2K$ probes of a text are issued as one batch. The window read takes the longest prefix whose prompt fits the window and records the truncation, for retrieval and for question answering alike. Every arm uses the same first-token letter readout, and no question is defaulted to a letter.

\textbf{Baselines.} BM25 uses $k_1{=}0.9$ and $b{=}0.4$. Qwen3-Embedding-4B uses last-token pooling with $L_2$ normalization and the standard retrieval instruction on the query side. For retrieval, documents are truncated to 2,048 tokens and scored by cosine similarity over the full corpus. For chunk selection on LongBench v2, every 500-token chunk is embedded and the top $K$ by cosine similarity to the question are read in original order. Random selection uses a fixed seed. Chain of Agents \citep{chainofagents} uses the worker and manager prompts of that paper's Table 9 verbatim: the workers read the $K$ local spans of the grid in order, the same spans as the local probes, each writing a summary of at most 512 tokens that carries the evidence to the next, and the manager answers from the last summary with the letter readout of every other arm, $K{+}1$ calls per question. Summaries are decoded greedily.

\subsection{Cost}\label{app:cost}

\textbf{Full cost with the prompt.} With a prompt of $p$ tokens, the single read costs $F_{\mathrm{full}}(s)=a(s+p)+b(s+p)^{2}$ and the $2K$ probes of $\sqrt{sc}+p$ tokens cost
\begin{equation*}
F_{\mathrm{probe}}(s)\;=\;2as+2b\sqrt{c}\,s^{3/2}+4bps+\bigl(2ap+2bp^{2}\bigr)\sqrt{s/c}.
\end{equation*}
The prompt terms are lower order in $s$, and the crossover of Eq.~\ref{eq:crossover} drops them.

Per-call compute is $2P_{\mathrm{ne}}t+2Ldt^{2}$ FLOPs for a prompt of $t$ tokens, with $P_{\mathrm{ne}}$ the non-embedding parameter count, $L$ the number of layers, and $d$ the hidden size. For Qwen3-4B-Instruct-2507, $P_{\mathrm{ne}}{=}3.633\times10^{9}$, $L{=}36$, $d{=}2560$, and the form matches hardware FLOP counters for single passes on this model. Token counts are measured on the exact prompt, template included. The FLOPs crossover of Section~\ref{sec:cost} is the positive root of $F_{\mathrm{full}}=F_{\mathrm{probe}}$, which with these coefficients and $c{=}100$ is 65.9k tokens for this dense model, and 15.7k against the local probes alone. In the notation of Section~\ref{sec:cost}, $a=2P_{\mathrm{ne}}=7.3\times10^{9}$ FLOPs per token, the attention coefficient $b=2Ld=1.8\times10^{5}$, and $p$ is the measured median prompt length per domain.

\textbf{Measured latency.} Figure~\ref{fig:cost}(a) in the main text is measured on one A100 80GB with vLLM, Qwen3.5-4B served at 1M tokens with YaRN, one single read against one batch of the $2K$ probes of the same text, random unique text so that prefix caching cannot help, both arms on the same idle server within the same minute. The single read takes 1.0, 1.9, 4.7, 12.9, 40, 137, and 457 s at 16k, 32k, 64k, 131k, 262k, 524k, and 996k tokens, the probes 1.7, 4.1, 6.7, 14.7, 26, 58, and 108 s in 12 to 90 calls. The probes cost more time up to 131k, 1.1 times there, and less from 262k on, 0.65 times at 262k and 0.24 at 1M. The longest single read this server completes is 996,220 tokens, the model length it was served at. Its cache would hold 2.1M.

Qwen3.5-4B and Qwen3.5-27B are hybrid models: 8 of 32 layers and 16 of 64 layers attend with full attention (hidden sizes 2,560 and 5,120, 16 and 24 attention heads of dimension 256, 4 key-value heads), and the remaining layers are linear in sequence length. The attention coefficient counts only the full-attention layers, $b = 2 L_{\mathrm{attn}} \times \mathrm{heads} \times 256$, which is $6.6\times10^{4}$ for the 4B and $2.0\times10^{5}$ for the 27B, against $a = 2P_{\mathrm{ne}}$ of $6.8\times10^{9}$ and $5.0\times10^{10}$. The crossover of Section~\ref{sec:cost} at $c{=}500$ is then 119k tokens for the 4B and 278k for the 27B. Below it the probes cost more than the single read, at 131k about the same for the 4B and 1.4 times for the 27B. Above it the ratio falls as $s^{-1/2}$: at 262k, the native window of these models, the probes cost 0.63 and 1.03 times the single read, at 1M tokens 0.23 and 0.44 times, and at the longest LongBench v2 context, 4.5M tokens, 0.07 and 0.13 times. On LongBench v2 the probes' compute tracks the token ratios of Section~\ref{sec:costexp}: 2.3 and 2.5 times the window read on the short and medium splits, and 14 times on the long split, where the window read stops at 131k tokens and the probes read the whole context. The exact prompt token counts of every call are stored with the results.

\textbf{Memory.} A full-attention layer stores keys and values for every token of a call, $2 \times 4 \times 256 \times 2$ bytes per token per layer with 4 key-value heads of dimension 256 in bfloat16, which is 64 KB per token over the 27B's 16 full-attention layers and 32 KB over the 4B's 8. The linear layers keep a state of fixed size. One read of the longest LongBench v2 context, 4,523,227 tokens, would therefore hold 296 GB of key-value cache for the 27B and 148 GB for the 4B. A probe holds at most $\sqrt{sc}$ tokens, 47.6k at that context, which is 3.1 GB and 1.6 GB. The serving engine's own accounting agrees: on one A100 80GB, vLLM allocates a cache of 2.1M tokens for the 4B and, next to the 27B's 54 GB of weights, 356k tokens for the 27B. A single read of the 27B past 356k tokens therefore does not exist on this hardware at any positional scaling, which covers 86 of the 108 long questions, and a single read of the 4B stops at 2.1M, short of 14 of them. The wall follows the budget: the same 4B served on the same card with vLLM capped to 16 GB holds 178k tokens (0.68 of the 262k it was configured for, in the engine's own report), completes reads of 131k and 160k tokens in 12.9 and 17.6 s, its longest completed read is 177,500 tokens and a read of 180,000 is never scheduled, and it still runs every probe, since none exceeds 47.6k tokens: the 46 probes of a 262k text finish in 24 s on that budget. At that budget the single read reaches 178k tokens and the probes $178\mathrm{k}^{2}/c = 63$M, the whole of LongBench v2 included. Reach grows with the square of the cache a device holds, which is what puts the method on small GPUs. A window in Section~\ref{sec:setup} is the context the server is configured for, not the point at which memory runs out. The point is that the probes' memory is bounded by $\sqrt{sc}$ whatever the context, and the single read's is not.

On BRIGHT most documents sit below the crossover and \method{} costs 2.6 to 27 times the FLOPs of the whole read by domain, a median of 5. The whole read fails on 0.7 to 26.1\% of documents by domain at the 16k window, and on 29\% overall in a companion run at a smaller window on 16GB-class hardware. The window is a property of budget and hardware and is reported with every such number.

\subsection{Per-Domain Retrieval Results}\label{app:domains}

\begin{table}[h]
\centering\small
\setlength{\tabcolsep}{4pt}
\caption{Per-domain NDCG@10, MRR, and Recall@10 on BRIGHT long-document retrieval, full corpus, 50 queries per domain. Bold: best per row.}
\label{tab:domains}
\begin{tabular}{@{} l cccccc @{}}
\toprule
\textbf{Domain} & BM25 & Embed & Embed, chunk-max & FirstP & Window read & \method{} \\
\midrule
\multicolumn{7}{@{}l}{\emph{NDCG@10}} \\
Biology & .206 & .565 & .514 & .508 & .622 & \textbf{.688} \\
Earth science & .184 & \textbf{.577} & .512 & .301 & .441 & .426 \\
Economics & .239 & .464 & \textbf{.471} & .263 & .441 & .455 \\
Pony & .355 & .143 & .069 & .099 & .273 & \textbf{.379} \\
Psychology & .184 & .561 & .551 & .160 & .532 & \textbf{.613} \\
Robotics & .146 & \textbf{.402} & .317 & .073 & .401 & .355 \\
Sustainable living & .220 & .513 & .521 & .152 & .463 & \textbf{.533} \\
Mean & .219 & .461 & .422 & .222 & .453 & \textbf{.493} \\
\midrule
\multicolumn{7}{@{}l}{\emph{MRR}} \\
Biology & .196 & .535 & .473 & .527 & .587 & \textbf{.675} \\
Earth science & .168 & \textbf{.579} & .492 & .273 & .426 & .396 \\
Economics & .228 & .388 & .401 & .253 & .385 & \textbf{.408} \\
Pony & .530 & .336 & .157 & .199 & .515 & \textbf{.576} \\
Psychology & .157 & .521 & .522 & .162 & .493 & \textbf{.597} \\
Robotics & .137 & .316 & .273 & .088 & \textbf{.349} & .298 \\
Sustainable living & .212 & .477 & \textbf{.489} & .181 & .417 & .485 \\
Mean & .233 & .450 & .401 & .240 & .453 & \textbf{.491} \\
\midrule
\multicolumn{7}{@{}l}{\emph{Recall@10}} \\
Biology & .318 & .830 & .760 & .637 & .803 & \textbf{.837} \\
Earth science & .323 & \textbf{.773} & .727 & .493 & .627 & .638 \\
Economics & .320 & \textbf{.740} & .720 & .350 & .660 & .660 \\
Pony & .394 & .145 & .079 & .122 & .286 & \textbf{.416} \\
Psychology & .341 & \textbf{.772} & .748 & .227 & .721 & .755 \\
Robotics & .230 & \textbf{.710} & .540 & .090 & .630 & .580 \\
Sustainable living & .298 & .730 & .732 & .184 & .706 & \textbf{.776} \\
Mean & .318 & \textbf{.672} & .615 & .300 & .633 & .666 \\
\bottomrule
\end{tabular}
\end{table}

Corpus sizes are 508 to 601 documents per domain. The map leads the window read on NDCG@10 and MRR in five domains and trails it in Earth science and Robotics, where the documents that fit the 16k window carry most of the relevance. Against the embedder the map leads MRR in five domains, Recall@1 in six, and Recall@10 in three: the verdict places a relevant document first more often, and similarity brings more of the several relevant documents into the top ten. The largest margin is on Pony, whose queries are code and where similarity to the query text is least informative.

\subsection{LongBench v2 Details}\label{app:paired}

\begin{table}[h]
\centering\small
\setlength{\tabcolsep}{4pt}
\caption{LongBench v2 accuracy (\%) by domain, all splits, both models. Bold: best per row.}
\label{tab:lbdomains}
\begin{tabular}{@{} l r ccc @{}}
\toprule
\textbf{Domain} & $n$ & Window read & Direct readout & Selected read \\
\midrule
\multicolumn{5}{@{}l}{\emph{Qwen3.5-4B}} \\
Single-document QA & 175 & 38.9 & 41.7 & \textbf{43.4} \\
Multi-document QA & 125 & \textbf{40.0} & 32.0 & 33.6 \\
Long in-context learning & 81 & \textbf{54.3} & 43.2 & 49.4 \\
Long-dialogue history & 39 & 43.6 & 51.3 & \textbf{53.8} \\
Code repository & 50 & 34.0 & \textbf{38.0} & 36.0 \\
Long structured data & 33 & \textbf{45.5} & \textbf{45.5} & 42.4 \\
\midrule
\multicolumn{5}{@{}l}{\emph{Qwen3.5-27B}} \\
Single-document QA & 175 & \textbf{50.9} & 48.6 & \textbf{50.9} \\
Multi-document QA & 125 & \textbf{53.6} & 46.4 & 44.8 \\
Long in-context learning & 81 & 59.3 & 63.0 & \textbf{64.2} \\
Long-dialogue history & 39 & 53.8 & \textbf{61.5} & 53.8 \\
Code repository & 50 & 54.0 & \textbf{58.0} & 56.0 \\
Long structured data & 33 & 51.5 & 54.5 & \textbf{57.6} \\
\bottomrule
\end{tabular}
\end{table}

Single-document QA and dialogue place the evidence in one region, and there the selected read leads or ties at both scales. In-context learning spreads its demonstrations through the text, and there the 4B trails the window read by 5 points and the 27B leads by 5. Structured data and code reward the document-wide view of the strided probes, and the 27B's beats the window read on both. Multi-document QA, whose evidence lies in different documents, is the one domain behind at both scales, by 6 and 9 points.

\begin{table}[h]
\centering\small
\caption{Questions won and lost against the 131k window read of Qwen3.5-4B on the same questions. $n{=}503$ unless stated.}
\label{tab:paired}
\begin{tabular}{@{} l r r r @{}}
\toprule
\textbf{Arm} & Wins & Losses & $n$ \\
\midrule
\method{}, selected read & 54 & 54 & 503 \\
\method{}, direct readout & 51 & 60 & 503 \\
First $K$ chunks & 39 & 93 & 503 \\
Random $K$ chunks & 44 & 89 & 503 \\
Embedding $K$ chunks & 51 & 82 & 503 \\
$K/2$ chunks & 52 & 63 & 503 \\
$2K$ chunks & 40 & 51 & 503 \\
$4K$ chunks & 40 & 46 & 492 \\
Local probes only, direct & 52 & 65 & 503 \\
Map $\to$ Qwen3.5-27B & 87 & 51 & 503 \\
Embedding $\to$ Qwen3.5-27B & 92 & 74 & 503 \\
Qwen3.5-27B window read & 106 & 48 & 503 \\
\midrule
\multicolumn{4}{@{}l}{\emph{Against the Qwen3.5-27B window read}} \\
Map from Qwen3.5-27B $\to$ Qwen3.5-27B, medium & 18 & 15 & 215 \\
Map from Qwen3.5-27B $\to$ Qwen3.5-27B, all & 43 & 47 & 503 \\
27B direct readout, all & 48 & 52 & 503 \\
27B direct readout, long & 13 & 5 & 108 \\
\midrule
\multicolumn{4}{@{}l}{\emph{Against the 4B reading 1M tokens with YaRN}} \\
\method{}, selected read & 58 & 44 & 503 \\
\method{}, direct readout & 54 & 49 & 503 \\
\bottomrule
\end{tabular}
\end{table}

\begin{table}[H]
\centering\small
\setlength{\tabcolsep}{4pt}
\caption{Accuracy (\%) by context length in tokens, both models, with bin edges at the 131k and 262k windows. Figure~\ref{fig:reach} (left) plots the 27B's native 262k read and both readouts. A window read past its window reads the longest prefix that fits.}
\label{tab:bins}
\begin{tabular}{@{} l l rrrrrr @{}}
\toprule
& & ${<}32$k & 32--64k & 64--131k & 131--262k & 262k--1M & ${>}1$M \\
\textbf{Model} & \textbf{Arm} & $n{=}111$ & $n{=}71$ & $n{=}116$ & $n{=}105$ & $n{=}67$ & $n{=}33$ \\
\midrule
Qwen3.5-4B & Window read, 32k & 43.2 & 39.4 & 29.3 & 37.1 & 32.8 & 36.4 \\
Qwen3.5-4B & Window read, 64k & 43.2 & 46.5 & 29.3 & 35.2 & 38.8 & 42.4 \\
Qwen3.5-4B & Window read, 131k & 44.1 & 43.7 & 34.5 & 43.8 & 44.8 & 45.5 \\
Qwen3.5-4B & Window read, 262k & 43.2 & 46.5 & 34.5 & 40.0 & 49.3 & 39.4 \\
Qwen3.5-4B & Window read, 1M (YaRN) & 41.4 & 45.1 & 29.3 & 41.0 & 40.3 & 45.5 \\
Qwen3.5-4B & \method{}, selected read & 45.9 & 32.4 & 39.7 & 42.9 & 49.3 & 39.4 \\
Qwen3.5-4B & \method{}, direct readout & 45.0 & 39.4 & 37.1 & 37.1 & 41.8 & 42.4 \\
\midrule
Qwen3.5-27B & Window read, 131k & 55.9 & 56.3 & 50.0 & 53.3 & 59.7 & 39.4 \\
Qwen3.5-27B & Window read, 262k & 56.8 & 56.3 & 49.1 & 58.1 & 64.2 & 51.5 \\
Qwen3.5-27B & \method{}, selected read & 54.1 & 47.9 & 53.4 & 49.5 & 56.7 & 57.6 \\
Qwen3.5-27B & \method{}, direct readout & 53.2 & 53.5 & 50.0 & 48.6 & 58.2 & 60.6 \\
\bottomrule
\end{tabular}
\end{table}

\begin{table}[H]
\centering\small
\setlength{\tabcolsep}{4pt}
\caption{Does the arm make the whole read's decision? The 298 LongBench v2 contexts the 131k window reads whole. \emph{Agree}: same answer as the whole read. $r$: Pearson correlation of the option log-probabilities with the whole read's, log-softmaxed over the four letters, 1,192 option points. $\rho$: mean per-question Spearman. The ceiling row is the whole read against a second run of itself in another session, on the 120 medium contexts read whole in both.}
\label{tab:fidelity}

\begin{tabular}{@{} l l rrrr @{}}
\toprule
\textbf{Model} & \textbf{Arm} & Agree (\%) & $r$ & $\rho$ & Acc. (\%) \\
\midrule
Qwen3.5-4B & \method{}, selected read & \textbf{70.5} & 0.82 & \textbf{0.70} & 40.3 \\
Qwen3.5-4B & \method{}, direct readout & 67.8 & \textbf{0.82} & 0.68 & \textbf{40.6} \\
Qwen3.5-4B & Embedding $K$ chunks & 64.4 & 0.76 & 0.60 & 36.2 \\
Qwen3.5-4B & Random $K$ chunks & 58.4 & 0.71 & 0.60 & 33.6 \\
Qwen3.5-4B & First $K$ chunks & 59.7 & 0.67 & 0.58 & 30.9 \\
Qwen3.5-4B & Whole read, re-run of itself (ceiling, $n{=}120$) & 95.0 & 1.00 & -- & -- \\
\midrule
Qwen3.5-27B & \method{}, selected read & \textbf{73.2} & \textbf{0.82} & \textbf{0.70} & \textbf{52.3} \\
Qwen3.5-27B & \method{}, direct readout & 71.1 & 0.80 & 0.67 & 52.0 \\
Qwen3.5-27B & Map of the 4B, $K$ chunks & 67.1 & 0.80 & 0.68 & 47.3 \\
Qwen3.5-27B & Embedding, $K$ chunks & 65.1 & 0.76 & 0.62 & 42.3 \\
\bottomrule
\end{tabular}
\end{table}

\begin{figure}[H]
\centering
\includegraphics[width=0.5\linewidth]{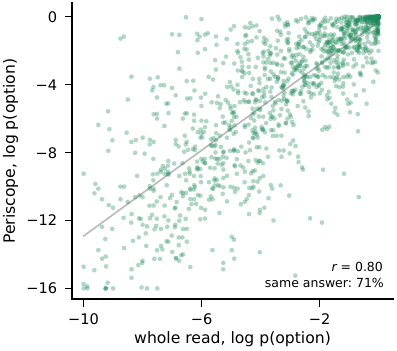}
\caption{Qwen3.5-27B, the direct readout's log-probability of each option against the whole read's, on the 298 contexts read whole.}
\label{fig:fidelity}
\end{figure}

A read that cannot fit the window is counted as unanswered and excluded from the accuracy rather than scored as wrong, which occurred for 11 questions at $4K$ chunks, whose accuracy is over the 492 answered, and for none of the arms in Table~\ref{tab:main}.

\textbf{Chunk size.} Doubling the chunk to 1,000 tokens cuts the number of probes by a factor of $\sqrt{2}$ and lengthens each by the same factor, and the selected read on the medium split falls from 41.9 to 38.6, so the map is sharper with the smaller chunk. This difference is close to the run-to-run variation above.

\begin{figure}[t]
\centering
\includegraphics[width=0.9\linewidth]{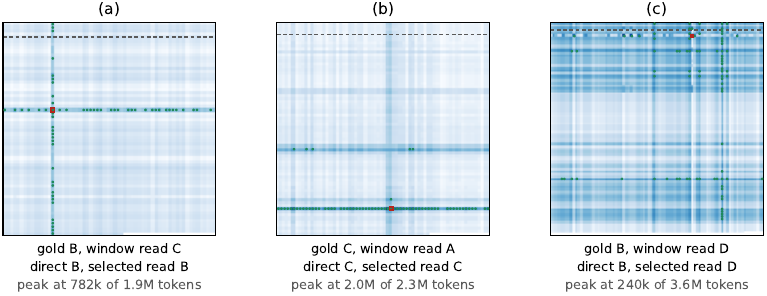}
\caption{What does the map look like? Three long-split questions, Qwen3.5-27B. Each horizontal line of cells is one local span and each vertical line one strided span, so the document reads line by line. The outlined cell is the peak, the dots are the $K$ selected chunks, and the dashed line marks the last local span a 131k read sees. (a) The peak's local and strided spans cross on the chunk that holds the answer, 782k tokens in, and the selection reads that local span. (b) The peak is 2.0M tokens into a price table, and the selection reads its local span. (c) A flat map, its peak 0.4 above the next cell: the selection scatters and the selected read misses, while the direct readout, which uses only the best score of each answer, still answers right.}
\label{fig:qual}
\end{figure}


\section{Chunk-Level Localization, Where the Document Fits in a Probe}\label{app:loc}

Two of the authors annotated 100 documents from the OBLIQ Congress corpus \citep{obliq}, marking for each the chunks containing evidence that justifies retrieval for the associated query and the most relevant among them, and a second annotation pass confirmed 53 of the 100. The annotations are anchored to character spans in each document, so the gold chunks are recomputed under the judge's tokenizer at scoring time. Documents are 10 to 30 chunks of 50 tokens, a median of 14 with about 4 annotated, so every metric is reported beside its value for a random order. Top-1 counts a query as localized when the peak of $M$ is an annotated chunk and MRR uses the rank of the first annotated chunk. Ties, which LocalOnly has inside every span, are scored by their expectation over random tie-breaks for every method. Every method scores the chunks of the given document against the query with the judge of Table~\ref{tab:bright}. The two whole reads are the judge reading the document once and pointing at a labeled passage, ranked by the log-probabilities of the labels, and the judge reading the document once per chunk with that chunk marked, $N$ reads.

\begin{table}[H]
\centering\small
\setlength{\tabcolsep}{5pt}
\caption{Does the peak land on the evidence when the whole document fits in one probe? 100 annotated OBLIQ documents, chunk level, given the document. Chance is a random order. \emph{Tokens}: judge tokens per query, in thousands.}
\label{tab:loc}
\begin{tabular}{@{} l ccc @{}}
\toprule
\textbf{Method} & Top-1 & MRR & Tokens \\
\midrule
Chance & 30.6 & 50.8 & -- \\
BM25 & 42.0 & 60.2 & -- \\
Qwen3-Embed-4B & 50.0 & 67.3 & -- \\
\midrule
FirstP & 9.0 & 26.7 & -- \\
FullRead, points once & \rc & \rc & 1.0 \\
FullRead, $N$ reads & 45.0 & 58.7 & 15.9 \\
LocalOnly & 47.9 & 63.9 & 1.5 \\
\method{} & 47.5 & 65.5 & 3.1 \\
\bottomrule
\end{tabular}
\end{table}

This is the boundary of the method, measured. Every document here is 500 to 1,500 tokens and fits inside one probe, so a strided span of a 4-chunk grid samples what the local span already covers, and there is nothing for the second family of probes to reach: the map and LocalOnly tie, and both tie the embedder. The reach of Eq.~\ref{eq:reach} is a statement about text longer than a call, and below that length the map is a retriever among retrievers. Two things hold even here. The map is above the judge's own whole read pointing at each chunk in turn, by 2.5 points of Top-1 and 7 of MRR at a fifth of the tokens, so composing bounded probes localizes better than the judge reading everything and pointing. And the map is 17 points above chance on Top-1, so the peak is evidence, not noise. On the 53 queries the second pass confirmed, every method moves by at most 5 points.


\end{document}